\documentclass[journal]{IEEEtran}
\usepackage{cite}
\usepackage{amsmath,amssymb,amsfonts}
\usepackage{graphicx}
\usepackage{textcomp}
\usepackage{algorithm,algorithmic}
\usepackage{xcolor}
\ifCLASSINFOpdf

\else

\fi

\begin{document}
\title{Batch Augmentation with Unimodal Fine-tuning for Multimodal Fusion of Large Language Models}

\author{
H M Dipu Kabir, Subrota Kumar Mondal,  Mohammad Ali Moni, \IEEEmembership{Member, IEEE}

\thanks{H M Dipu Kabir and Mohammad Ali Moni are with AI and Cyber Futures Institute, Charles Sturt University, Australia}
\thanks{H M Dipu Kabir and Mohammad Ali Moni are also with Rural Health Research Institute, Charles Sturt University, Australia.}
\thanks{Subrota Kumar Mondal is with the School of Computer Science and Engineering, Macau University of Science and Technology, Macao.}

}

\maketitle

\begin{abstract}
In this paper, we propose batch augmentation with unimodal fine-tuning for multimodal learning. We start with pre-trained unimodal models. We fine-tune the unimodal models with the application data. After that, we form a Multi-Layer Perceptron (MLP) head that takes information from unimodal models and provides output. Finally, we train the MLP layer and unimodal parts with batch augmentation. 
Depending on the data, some unimodal models can be replaced by hard-coded scripts or AI agents. The unimodal training can also follow batch augmentation when the data is augmentable.
We write a multimodal batch augmentation dataloader script that implements the batch augmentation for the multimodal data. We investigate the proposed method on the FPU23 ultrasound and UPMC Food-101 multimodal datasets. The multimodal large language model (LLM) with the proposed training achieves the best average result among the investigated methods across both datasets. According to our literature search, the proposed method achieves state-of-the-art (SOTA) accuracy of 93.29\% on the UPMC Food-101 dataset, while we apply the ViT-L/16 model for vision and the GPT-2 model for text. We share the scripts of the proposed method with traditional counterparts at the following repository: github.com/dipuk0506/multimodal

\end{abstract}

\begin{IEEEkeywords}
LLM, Ultrasound, Transferred Initialization, Multimodal Learning, Dataloader.
\end{IEEEkeywords}

\IEEEpeerreviewmaketitle

\section{Introduction}
\label{sec:introduction}
\IEEEPARstart{D}{ata} augmentation is a popular technique for improving generalization. The augmentation increases both the count and diversity of data \cite{hoffer2020augment}. The initial dataset may contain a few patterns. Especially, in the medical domain, it is difficult to find many patients with rare diseases. While applying the model to patients, the collected sample can be slightly different from the training samples. However, there exist several common patterns in both images. When the samples are images, the test image can be a shifted, rotated, blurry, or any other version of the training image. Several researchers considered feature extraction followed by another model training \cite{chen2020image, sun2020dl}. However, data loading and saving require more time than computation. Moreover, image data augmentation increases the number of samples by thousands of times \cite{wang2017effectiveness}. Saving all features with all possible combinations requires a lot of memory. Moreover, loading images from different random locations to achieve a varying augmentation in a batch significantly increases the data loading time \cite{manegold2002generic}. Recent random augmentation functions from the TorchVision library with Dataloader and training help us to augment different images in a batch differently \cite{marcel2010torchvision, albardi2021comprehensive}. Therefore, the optimization becomes robust while the scheduler steps on each batch.

\begin{figure*}
\begin{center}
\centerline{\includegraphics[clip, trim=1.0cm 12.2cm 22.2cm 21.4cm, width=5.0in,angle=0]{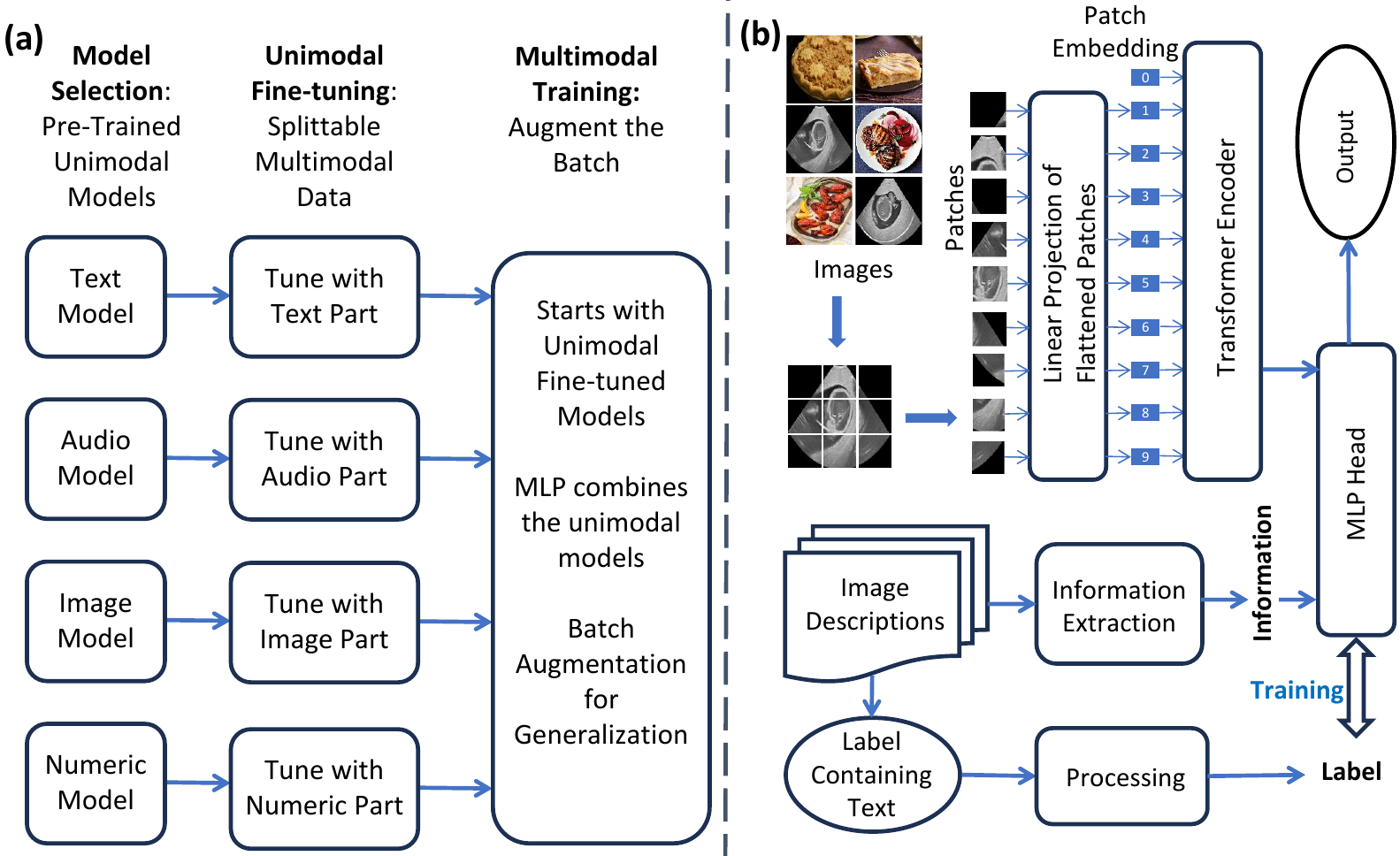}}
\caption{The proposed training and implementation. (a) A rough sketch of the proposed multimodal training with unimodal fine-tuning and batch augmentation. (b)Information flow in the proposed multimodal learning during implementation. We extract features from images through pre-trained initial layers. We extract information and labels from text. This figure shows a vision transformer (ViT) to obtain image features. We also apply a ResNet-type model for feature extraction and a GPT-2 model for text processing. A Multi-Layer Perceptron (MLP) head combines the unimodal models.}
\label{MM_tr}
\end{center}
\end{figure*}

Ultrasound imaging is one of the popular techniques to observe the growth and health of the fetus. It is the best method to observe the fetus regarding safety and cost-effectiveness. X-ray imaging and CT (Computed Tomography) scans apply ionizing radiation \cite{
abuelhia2020evaluation}. MRI (Magnetic Resonance Imaging) does not use radiation. However, it uses strong magnetic fields. MRI is performed using expensive machines compared to ultrasound \cite{bekiesinska2012fetal}. Although ultrasound does not apply harmful radiation or strong magnetic fields, the resolution of ultrasound images is much lower than CT scan images. Ultrasound machines are usually cheaper than X-ray and CT scans \cite{gibbons2021lung}. As a result, many hospitals and diagnostic centers in less developed countries are using ultrasound to observe the condition of the fetus. There is a lack of efficient people to take ultrasound images. As a result, ultrasound images collected in those locations are usually noisier. It is hard for people and machine learning algorithms to detect fetal organs from ultrasound images \cite{stewart2020trends}.

There is a lack of medical imaging specialists in underdeveloped countries and remote areas of developed countries \cite{shaddock2022potential, burgos2020evaluation}. An initial screening of the medical image with the help of Artificial Intelligence (AI) can potentially reduce the load of experts. When the machine detects an organ or provides an initial report, the doctor can potentially decide quickly and with less effort. Moreover, the machine can refer images to different medical practitioners based on the prediction and associated uncertainties \cite{kabir2022aleatory}.

Fig. \ref{MM_tr} presents the proposed training steps and example information flow of the proposed multimodal network during execution. The proposed multimodal learning first fine-tunes unimodal models before a combined multimodal training with batch augmentation. Unimodal models are joined by a multi-layer perceptron (MLP) head. We used a pre-trained vision transformer (ViT) from the \emph{timm} library \cite{wightman2021resnet} and a GPT-2 model from OpenAI \cite{roumeliotis2023chatgpt} to investigate our proposal. We also extract information from texts and generate numeric values. Except for the label, we provide those numeric values as inputs along with features to the \emph{Head Layer} of the model.

The learning capability of humans is limited. Humans can learn from thousands of samples. However, artificial intelligence (AI) models can learn from billions of samples \cite{wu2022robust}. Therefore, recent state-of-the-art (SOTA) performing AIs are better than humans in recognizing natural images. Humans have both genetic inclination and training from childhood to recognize natural images. However, AI has overpowered humans in recognizing natural images. Medical images are not usually natural images \cite{greenspan2016guest}. We do not find ultrasound, X-ray, or CT scan images in nature. They are generated through machines. 
A human doctor can potentially learn and memorize from thousands of images. Fortunately, humans can relate descriptions and other information with images to make a decision. Therefore, integrating other information with medical images and model training can potentially bring better diagnoses than human doctors in the future. 

According to our literature search and theoretical understanding, contributions in this paper are as follows:
\begin{enumerate}
    \item According to our literature search, we propose batch augmentation for multimodal medical data for the first time.
   \item We share a data loader script that can use both custom augmentation and standard augmentations from the PyTorch library.
  \item We integrate features from neural networks (NNs), large language models (LLMs), and logical screening.
  \item We convert a vision transformer to a multimodal image-text model for fetus organ detection for the first time.
  \item We also fuse a vision transformer (ViT-L/16) with a language transformer (GPT-2).
  \item We also prescribe the further training of initial layers with current data before the multimodal training when the current data contains quite different features compared to the dataset of pre-training.
\end{enumerate}

\section{Background and Related Works}
\label{Rworks}
This section presents several theories and other information related to the proposed method to help readers.

\subsection{Feature Propagation in Transfer-Learned Models}
Initial layers of deep classification NNs create features, and end layers compute scores for different classes from those features. The concept of class activation map showed that the spatial position ($x,y$) in the last convolutional layer is directly linked to the same relative spatial position in the input image \cite{zhou2016learning}. Therefore, the feature at ($x,y$) position in the last convolutional layer comes from the same relative spatial position ($x_i,y_i$) on the image. The following equation computes the score for \emph{c} class: 
\begin{equation}
S_c = \sum_{x,y} \sum_{k} w_k^c(x,y) Au(k,x,y),
        \label{CAM1}
\end{equation}
where, $Au(k,x,y)$ is the $k^{th}$ activation unit at $(x,y)$ position and $w_k^c(x,y)$ is the weight connection between the class ($c$) output and $Au(k,x,y)$.

The initial layers of convolutional NNs compute low-level features from images. Features in deeper layers contain high-level and more output-related information. The first few initial layers of CNN compute textures, corners, edges, etc. Mid-layers of CNN compute shapes using outputs of previous layers. Deep layers of CNN contain information on the part of the classification object. The weights on deeper layers are dependent on the target classification problem \cite{simonyan2013deep, yosinski2015understanding}. When the classification problem is classifying images of animals, the deeper layers contain patterns available on images of animals \cite{zeiler2014visualizing}. Although several recent transformer-type models do not contain convolutional parts, they segment images into patches and obtain high-level features from patches over layers. Finally, a fully connected head layer decodes the outputs of the transformer encoder into classification scores. Deeper layers have deeper biases on the pre-trained dataset.

Medical images are quite different compared to the natural images. When a deep NN is pre-trained on natural images, deep layers become good at identifying natural patterns. Patterns in medical images are quite different compared to natural images. Deep convolutional layers of that NN may not propagate all important patterns to the head layer. As a result, the classification accuracy of NN on the medical dataset becomes low. 

Although the vanishing gradient is a problem while training deep NN from scratch, it becomes a blessing when researchers train a deep NN with transferred initialization \cite{kabir2022spinalnet}. The fully connected end layers get a different size with random initialization based on the target dataset. Deeper layers are either randomly initialized or contain high bias on the dataset of pre-training. Therefore, deeper layers need more change in the values of their weights. Initial layers perform basic operations that do not vary significantly from dataset to dataset. Therefore, values of initial layer weights need very slight or no change. Due to the vanishing gradient, the initial layer weights remain almost the same, and the end layers undergo drastic changes.

\subsection{Batch Augmentation}
Different augmentation of different samples in a batch brings a better generalization \cite{hoffer2020augment}. Training of convolutional neural network (CNN) on the Modified National Institute of Standards and Technology database (MNIST database) without augmentation brings about 98.50\% accuracy. While training with random perspective and random rotation brings about 99.50\% accuracy \cite{kabir2022spinalnet}. Although it seems that the accuracy is reduced by about 1\%, the error is becoming one-third. Augmentation brings a significant improvement in the performance of the machine learning model.

Neural Network (NN) training methods usually load datasets in batches and compute errors for each batch due to memory limitations. The optimizer steps based on the error. The optimizer step updates the weights of the NN model. The loss function can be expressed as, $l(f(),x_n, y_n),$ where $f()$ is the model, $x_n$ is the example input, and $y_n$ is the example output. $f()$ contains weights ($w$) and the model structure organizing weights. The updated weight for the $i+1$ iteration becomes as follows: 
\begin{equation}
w_{i+1} = w_i - \frac{\eta}{B_N} \sum_{n=1}^{B_N} \Delta_{f()} l\Big(f()_i,x_n, y_n\Big),
\end{equation}
where $\eta$ is the learning rate and $B_N$ is the number of samples on the batch. In a training, validation, or test phase, all batches contain the same number of samples except the last batch. The last batch in a phase can contain a fewer sample when the remaining number of samples for the last batch is lower than $B_N$. A significant portion of transfer learning and multimodal learning papers in the medical domain use no augmentation \cite{albaqami2021automatic, jin2020deep}. They apply initial layers of NNs to images without augmentations to obtain features. After that, they train a fully connected NN head on the features. 

When there is a constant augmentation $Aug_c()$ for inputs $(x_n)$ in a batch, the updated weight for the $i+1$ iteration becomes as follows: 
\begin{equation}
w_{i+1} = w_i - \frac{\eta}{B_N} \sum_{n=1}^{B_N} \Delta_{f()} l\Big(f()_i,Aug_c(x_n), y_n\Big).
\end{equation}
When the augmentation is constant over $n$, all the samples in a batch receive the same augmentation ($Aug_c()$). As a result, all the samples in a batch get the same adversity, and the model learns to tackle a constant adversity over the computation on the batch. The trained model faces the following limitations in such situations:
\begin{enumerate}
    \item The weight update ($w_{i+1}$) over a batch can significantly degrade the performance of NN on usual images.
   \item The NN becomes robust against one type of adversity $Aug_c()$ during the update. However, the performance of NN in other kinds of adversity can significantly degrade.
\end{enumerate}
These limitations can be minimized with a small batch size with a low learning rate \cite{kandel2020effect}. A small batch size is not feasible for large multimodal data. A small batch size keeps unused resources and makes the training time longer. Therefore, a random augmentation function that can select different augmentation transformations randomly for different images can reduce the computation cost and bring generalization at a lower computational overhead. The weight update for the random augmentation is as follows:
\begin{equation}
w_{i+1} = w_i - \frac{\eta}{B_N} \sum_{n=1}^{B_N} \Delta_{f()} l\Big(f()_i,Aug_v(x_n,n), y_n\Big),
\end{equation}
where, $Aug_v(x_n,n)$ is the variable augmentation function. This augmentation function varies over the sample number ($n$). There exist several library functions for random augmentation. When we call a random augmentation of ten degrees, the augmented image can get any rotation between negative ten degrees and positive ten degrees. Some of the images in a batch may not receive any augmentation for certain values of random variables.  

\begin{figure}
\begin{center}
\includegraphics[clip, trim=1.2cm 6.8cm 8.7cm 3.4cm, width=3.5in,angle=0]{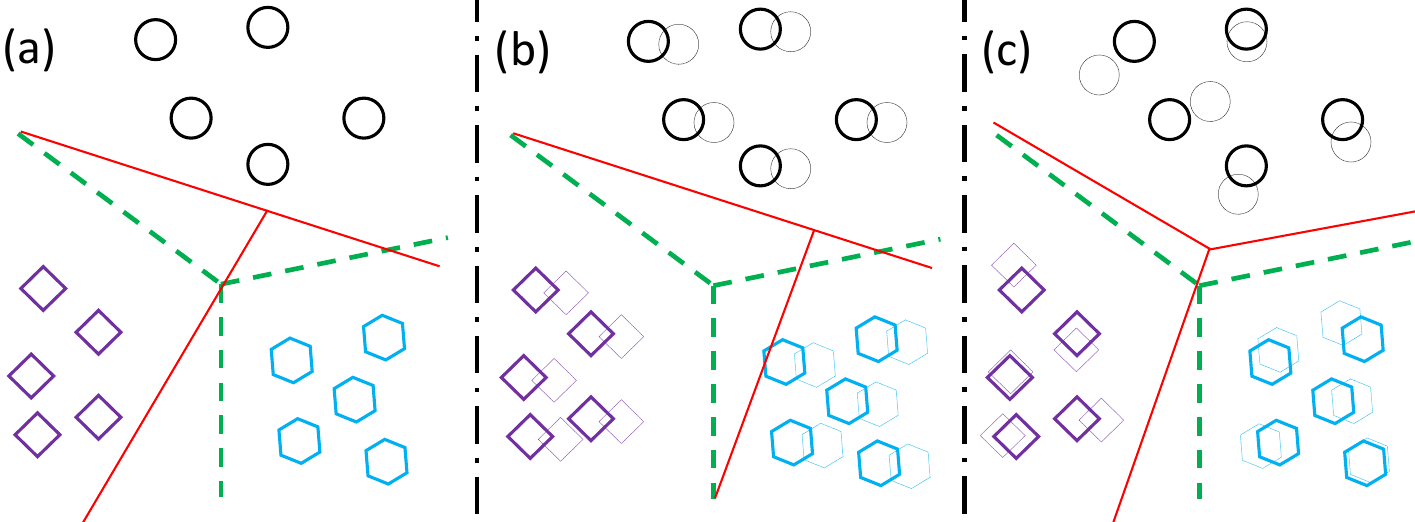}
\caption{Visualization of the importance of batch augmentation. Rough diagram (a) presents samples of different classes with different shapes. The green dotted lines represent the ground truth decision boundary. Red solid lines show the decision boundary of a poorly trained NN. Rough diagram (b) presents the effect while all samples in a batch receive the same augmentation. Rough diagram (c) presents the situation while different samples receive different augmentations. Patterns with thick solid lines present original samples and patterns with thin dotted lines present augmented samples.}\label{Batch_Aug}
\end{center}
\end{figure}

Fig. \ref{Batch_Aug} presents the importance of batch augmentation. The rough diagram in Fig. \ref{Batch_Aug}(a) presents samples of different classes with different shapes. The green dotted lines represent the ground truth decision boundary. Red solid lines show the decision boundary of a poorly trained NN. Rough diagram Fig. \ref{Batch_Aug}(b) presents the effect while all samples in a batch receive the same augmentation. When all samples in a batch receive the same augmentation, all samples shift in the same direction on the input domain. The decision boundary of the NN also shifts. Therefore, the NN does not become robust enough against other augmentations. Moreover, the trained NN may fail to predict many usual samples due to the shift of decision boundary. Rough diagram Fig. \ref{Batch_Aug}(c) presents the situation while different samples receive different augmentations. When the batch contains many samples and samples are augmented randomly the decision boundaries come closer to the ground truth decision boundary.



\subsection{Importance of Fetal Imaging and Organ Detection}
Fetal imaging is linked to complex culture and politics of reproduction \cite{petchesky2021fetal}. The allowance for abortion is a quite debated decision in the United States. The imaging of the fetus helped them understand or realize many concerns. Even today, different states have different policies regarding the validity and fetus age limit for abortion \cite{vilda2021state}. Moreover, the feasibility of abortion depends on the availability of facilities and insurance policies. Social perspectives and many policies on abortion are also linked to maternal health and maternal mortality \cite{vilda2021state}.

High-frequency sound waves are applied for imaging in the ultrasound imaging method. Several recent machines are providing 3D ultrasound. For example, Philips developed GlassVue, and Samsung developed CrystalVue. These machines use advanced software for better visualization of the fetus. MRI is usually used for more detailed observation of the fetus due to its high resolution. Fluorescence Microscopy is usually used to detect cancer cells. It is used in observing embryos. Fluorescence Microscopy can be used to observe the veins of the fetus.  However, this technique is used to observe an embryo when it is outside of the mother's womb. CT scan fetal imaging is not popular due to its high cost and safety concerns. 
Although there exist several popular imaging techniques, ultrasound remains the main fetal imaging means due to its portability, affordability, and safety. 


AI has been used in numerous medical data including the ultrasound data \cite{bayisa2024unified}.  
Several researchers have applied machine learning models to ultrasound data for observing the fetus \cite{horgan2023artificial}. Gofer et al. tried to segment and classify brain images of fetuses \cite{gofer2022machine} using AI models. Tsai et al. \cite{tsai2020automatic} and Nie et al. \cite{nie2017automatic} developed machine learning methods to predict the fetal plane. Several researchers also performed placental studies \cite{gupta2022ultrasound}. Researchers also worked on fetal biometry prediction \cite{arroyo2022no, prieto2021automated}. Several works exist on fetal heart monitoring \cite{sakai2022medical}. In this study, we have applied AI models for several organ detections. We apply multimodal learning with random augmentation in each batch for the fetus organ detection for the first time. Moreover, we have used a larger dataset compared to most other studies to observe the improvement brought by the proposed method. 


\subsection{Multimodal Learning}
Deep learning has shown promising performance in numerous areas. With the advancement of deep learning, machines can perform myriad tasks; that only humans could perform. However, most machine-learning approaches use data in one format. Some machine learning models are trained on image data, while some other models are trained on text data, and some other models are trained on audio signals. Multimodal learning is a branch of machine learning where models are trained on two or more different types of data. While considering more information from multiple sources, the system becomes more robust against failures. Deep multimodal learning is getting more and more popular over time due to its emerging needs \cite{ramachandram2017deep}. 

The act of a multimodal network is quite similar to tasting foods. When we see any food, our brain detects the food. Sometimes the image is not enough to judge the quality of food. To ensure the quality, we grab the food. Our touch sensory detects the hardness and texture of food. When we buy fruits, we often check the hardness and texture to ensure the quality. We also smell foods to check their condition. When we are biting food, our teeth provide predictions on the hardness of the food.  Our tongue performs a complex prediction on the cooking and ingredients. When we are swallowing, our oesophagus provides us feedback when there is any tiny sharp ingredient in the food. The absence of one or more of these sensory organs and their prediction can potentially lead us to eat the wrong food. Besides sensory organs, humans can eat poisonous food due to a lack of knowledge. Data processing from multiple sources reduces the chance of making errors.

Multimodal learning has become vital in many areas \cite{ramachandram2017deep, ngiam2011multimodal}. Humans show quite complex behavior in society. Social scientists get data on human actions from various sources and apply multimodal learning for the prediction. A multimodal system is a must for developing an autonomous system. Fully autonomous driving requires map information, images collected from cameras, LiDARs, GPS, ultrasonic sensors, etc. \cite{cui2024survey}. The fusion is a must for the development of autonomous systems. Doctors observe medical images, numerical information, patients' tones, statements, and body movements to make a decision. Therefore, researchers in medical domains also need multimodal AI models to make more accurate medical diagnoses.

\section{Proposed Method}
\label{proposed}
This section presents the proposed methodology with the help of theory and data. Therefore, we present datasets at first. After that, we explain methods with the help of data.
\subsection{Investigated Datasets}
We have investigated our proposed method on the FPU23 dataset \cite{prabakaran2023fpus23} and multomodal Food-101 datset \cite{wang2015recipe}.

\begin{figure}
\begin{center}
\includegraphics[clip, trim=2.6cm 9.6cm 3.1cm 1.3cm, width=3.4in,angle=0]{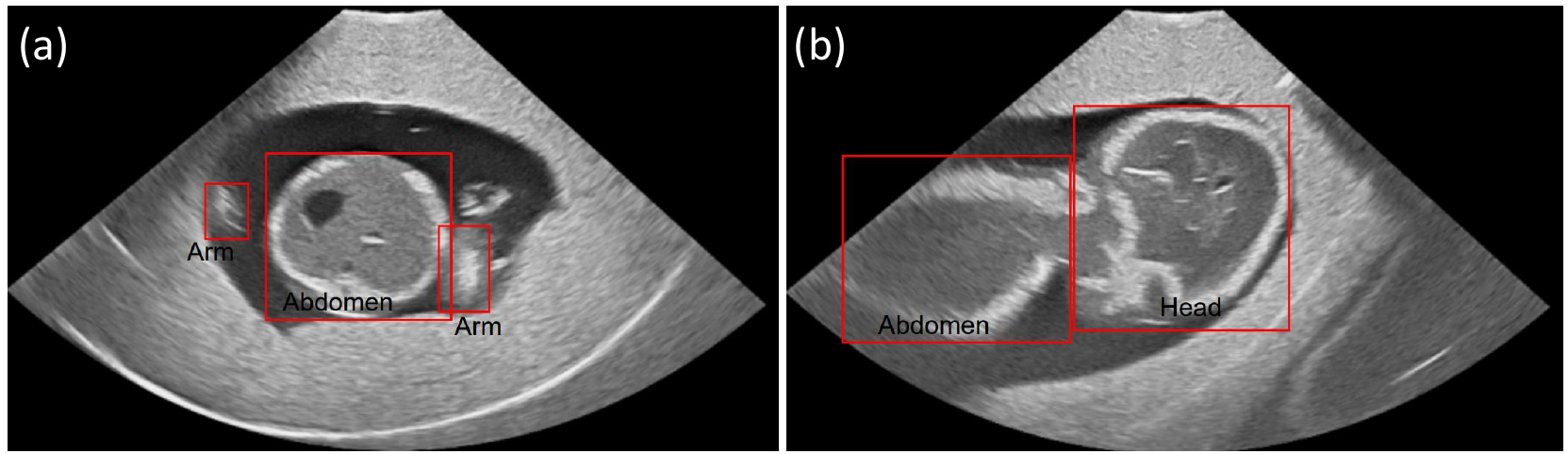}
\caption{Two example image on the FPU23 dataset. (a) The image contains the abdomen and two arms. (b) The image contains the head and the abdomen. }\label{Images2}
\end{center}
\end{figure}

The FPU23 dataset contains ultrasound images taken from different methods and for various positions and orientations of the fetus. Fig. \ref{Images2} presents two representative images of the dataset with labels. Fig. \ref{Images2}(a) contains the abdomen and arms. Fig. \ref{Images2}(b) contains the head and the abdomen. Both the presence and position of the organs are available with the dataset. Description of the position and the orientation of features are available as texts on an \emph{.xaml} file. A description of the image collection method is also available as text. Annotation boxes are not part of images. The presence of organs and positions annotation boxes are also available as texts.

The dataset contains more than fifteen thousand images. We split the dataset into training, validation, and test subsets. Ultrasound images were collected at the fetal ages of twenty-three weeks. Providers of the dataset also trained initial models to detect the orientations of the fetus, diagnostic planes, etc. We have extracted different texts from image descriptions and loaded images.

Table \ref{TABComp1} presents the number of images for different fetus orientations and data collection combinations. According to the table, the dataset is almost uniformly distributed into various combinations. There is a slight difference. There are slightly fewer samples with the invasive approach than the other. The number of head-up fetus images is slightly higher than the number of head-down combinations.

\begin{table} 
		\caption{Distribution of the FPU23 Data.}
		\label{TABComp1}
		\centering
			\begin{tabular}{|c|c|c|c|c|}	\hline
    head up      &view front      & Image       & Number\\
    (hu) or head &(vf) or view    & Collection  & of    \\
    down (hd)    &back (vb)       &(Invasive?) 	& Images\\ \hline
    hu           &vb              &Yes          & 1655 \\	\hline
    hu           &vb              &No           & 2021 \\ \hline
    hu           &vf              &Yes          & 2185 \\	\hline
    hu           &vf              &No           & 2571 \\ \hline
    
    hd           &vb              &Yes          & 1842 \\ \hline 
    hd           &vb              &No           & 1604 \\ \hline 
    hd           &vf              &Yes          & 1513 \\ \hline 
    hd           &vf              &No           & 1722 \\ \hline 
	    \end{tabular} 
\end{table}	

\begin{figure}
\begin{center}
\includegraphics[clip, trim=6.8cm 6.3cm 2.1cm 5.6cm, width=2.0in,angle=0]{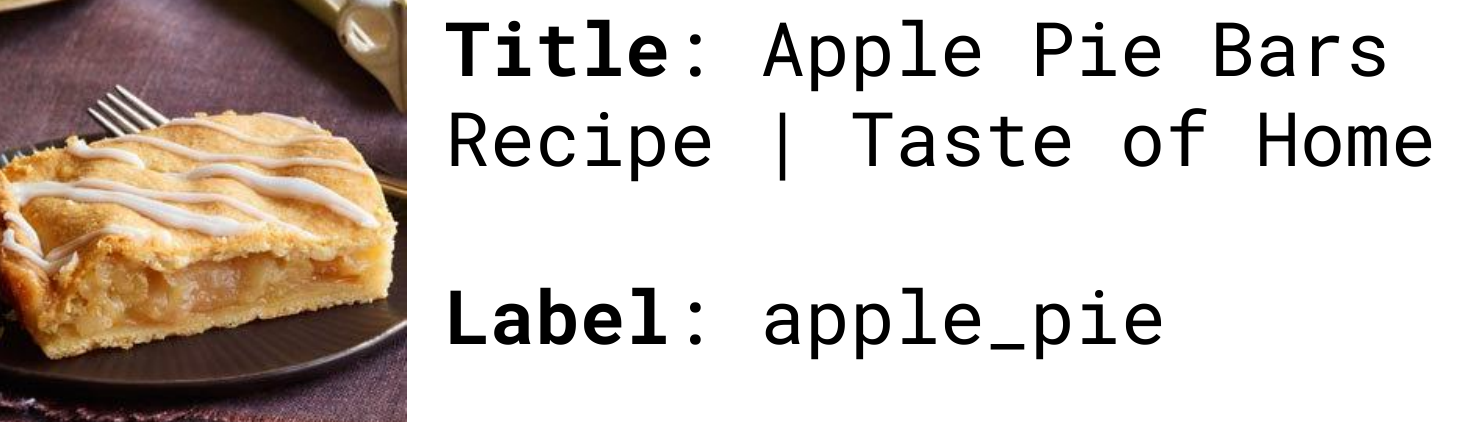}
\caption{An example sample on the UPMC Food-101 dataset. Each sample contains an image, a title, and a label. }\label{Apple_pie}
\end{center}
\end{figure}

Researchers at the University Pierre and Marie Curie (UPMC) developed a multimodal version of the Food-101 dataset \cite{wang2015recipe}. This dataset is also known as the UPMC Food-101 dataset. The dataset contains more than a hundred thousand food images categorized into 101 classes. The training dataset contains about sixty-eight thousand images and the test dataset contains about twenty-eight thousand images. The dataset contains images, titles, and labels for each sample. Fig. \ref{Apple_pie} presents an example sample on the dataset. The image contains an image of an apple pie slice with a title and label. The title is ``Mom's Maple-Apple Pie Recipe $|$ Taste of Home" and the label for this sample is `apple\_pie.'

\subsection{Pre-processing of Image Data}
\begin{figure}
\begin{center}
\includegraphics[width=3.4in]{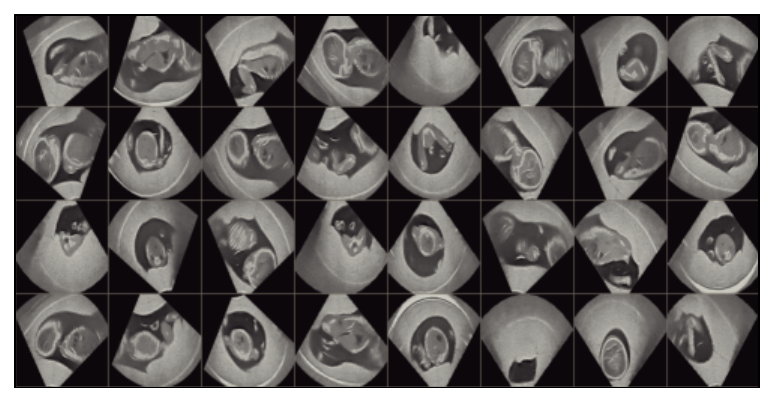}
\caption{Visualization of all images in a batch of 32 images after augmentation on the FPU23 dataset. We apply random rotations, random crops, random horizontal flips, and random vertical flips augmentations on training images.}\label{augment}
\end{center}
\end{figure}

\begin{figure}
\begin{center}
\includegraphics[width=3.4in]{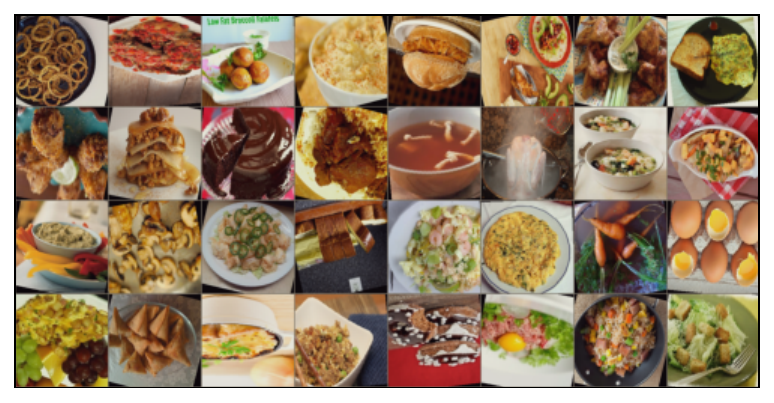}
\caption{Visualization of all images in a batch of 32 images after augmentation on the UPMC Food-101 dataset. We apply random rotations, random crops, random horizontal flips, and random vertical flips augmentations on training images.}\label{food}
\end{center}
\end{figure}

The proposed data pre-processing technique consists of several steps. We pre-process both texts and images. We assign numeric values for different image conditions. Also, we augment images for proper generalization. At the first step of image pre-processing, we resize images to 244 by 244 size. After that, we perform a random rotation of fifteen degrees. After that, we crop images of 224 by 224 size. After that, we apply random horizontal and random vertical flips. Finally, we normalize images. As TorchVision provides standard image augmentation functions, we used them instead of making customized augmentation functions. Fig. \ref{augment} presents a batch of images on the FPU23 dataset after augmentation. Fig. \ref{food} presents a batch of images on the UPMC Food-101 dataset after augmentation. The batch size is 32 in these images. We use a larger batch for \emph{ResNet-50} and a smaller batch for the \emph{ViT-L/16} model training to optimize training time considering available GPU memory.

\subsection{Pre-processing of Text and Numeric Data}
The UPMC Food-101 data contains text in natural human language. Such text data needs to be tokenized. When there is no existing vocabulary, the model developer may need to build a vocabulary from the training dataset. Moreover, the developer needs to handle unexpected words or tokens on the test set. Both recurrent NN and traditional shallow NN can be trained to obtain scores for different classes. 

GPT-2 model is one of the most popular and freely trainable text processing models \cite{zhang2023survey}. Pre-trained GPT-2 can provide promising performance, and it does not require developing a vocabulary. As GPT-2 is trained on a large dataset, it has a large collection of words in its vocabulary. However, the model needs a tokenizer. The fine-tuned GPT-2 model also provides scores for different classes.

We concatenate scores of different classes with features of the image network to feed the multimodal fully connected or head layer. The FPU23 dataset contains the collection method as extra information. We convert that information to numbers and concatenate those numbers with features of the image network to feed the multimodal fully connected or head layer.

\subsection{The Multimodal Framework}
Fig. \ref{MM_tr} presents the information flow in the proposed multimodal framework. Investigated two datasets that have slightly different data structures. Both datasets contain images, and the text containing files contains information on image links.

In the FPU23 dataset, we read the \emph{.xaml} (Extensible Application Markup Language) files of the dataset and extract information. The proposed method takes the names of images, their labels, and other information from the description. Pre-trained initial layers of a model extract feature from images. Moreover, relevant information is extracted, and numbers are assigned to each combination as the Head layer takes only numbers. 

In the UPMC Food-101 dataset, the text files contain image names, labels, and titles of samples. We train a shallow NN and the GPT-2 model to get scores for all the classes. We consider these scores as extra features of the multimodal model. 

We normalize the extracted numeric information. We provide the normalized information to the head layer. The label of each image is also extracted from the text. This figure presents one vision transformer. Vision transformers are the recent SOTA-performing models. We also demonstrate the proposed method with the \emph{ResNet-50} model. The proposed multimodal learning can potentially be applied to any model.

\subsection{Multimodal Dataloader}
We wrote the script of the Dataloader based on the dataset and the requirement of the proposed multimodal training. The Dataloader loads images, texts containing labels, and other image descriptions. We wrote a robust Dataloader for all common types of fetal organ detection. The Dataloader finds the presence of the label by searching the word in the text. The Dataloader also processes several texts containing the orientation of the fetus, the sample collection process, and the direction of imaging. The Dataloader converts that information to unique numbers. Many images contain multiple labels. Therefore, we keep all labels as texts. Dataloader loads the text and searches for the presence of the label based on the detection problem.
 
Researchers are computing features from the entire dataset and saving the features in many medical image-processing papers. Later, they concatenate features and train the head layer based on the concatenated feature \cite{huynh2016digital, kieffer2017convolutional, alhindi2018comparing}.  However, it is not feasible to save all features with all possible augmentations. When the dataset contains several thousand features, the number of possible augmented images becomes several million. The space we need to save all image features can be a thousand times the size of the dataset. Loading data requires more time in computation machines than computing. Moreover, different images in the same batch need different augmentations for a good generalization. Current SOTA-performing methods are also using random augmentations \cite{yu2022coca, kabir2022spinalnet}. Therefore, the Dataloader augments different images differently using the library augmentation functions from TorchVision \cite{marcel2010torchvision}. 

\subsection{Proposed Multimodal Training}
Algorithm \ref{ALG_Train} presents the proposed multimodal training method. We start with a pre-trained model. We collect features from the initial layers of models. We concatenate those features with other extracted information in the later step. Therefore, we omit the head layer of the model and then apply the model to compute features from images. We create a head layer $NN_H$ based on the designed number of inputs and outputs. We set the detection variable `D' based on the detection problem. The detection variable is set to `head', `abdomen', `arm', and `legs' respectively for head, abdomen, arm, and leg detections. In the image information of the dataset, the `arm' word is used for the presence of an arm, and the `legs' word is used for the presence of a leg on the image. There can be slight differences in the process based on the organization of data. In the FPU23 data, we need to extract labels based on the problem. However, each sample has a single label on the Food-101 data. The other information in the FPU23 data is the plane combinations. Each combination gets a number as the extra information, obtained in line 10 of Algorithm \ref{ALG_Train}. However, the UPMC Food-101 dataset contains titles. We train a separate neural network to obtain features from texts.

We train the head layer in training sessions. We assign zero values to the best accuracy. We also save the weights of the model as the initial best. If the model can classify a few validation images in the first epoch, the accuracy of the model on that epoch becomes greater than zero. That non-zero accuracy becomes the new best accuracy and the weights of the model after that epoch become the weights of the new best model.  
We initialize the optimizer and the scheduler with the declared head layer ($NN_H$), learning rate, momentum, step size, and gamma values. We load images with the Dataloader and apply augmentations to images. 
 
In each training epoch, we run two phases: the `Training' phase and the `Validation' phase. The data is loaded on each phase in batches. The size of the batch depends on the capability of the machine. The batch size is sixty-four for the training of the \emph{ResNet-50} model. The batch size is twenty for the training of the \emph{ViT-L/16} model. We extract features in batches from images ($F_{Img}[j]$) using the initial layers of pre-trained NNs. We extract information ($Info[j]$) from the text in each batch. We normalize $Info[j]$ and concatenate with $F_{Img}[j]$. We apply the newly declared head layer ($NN_H$) to concatenated information and obtain the prediction ($P_H[i,j]$). We compute loss ($Loss[i,j]$) and accuracy ($Acc[i,j]$) values from predictions and labels.  
 
When the phase is `Training', we perform the optimization step using the loss function and optimizer. When the phase is `Validation', we save accuracy with the population count for each batch. These accuracies are used to compute the overall validation accuracies at an epoch. If the overall validation accuracy in an epoch is higher than all previously recorded accuracies, the best accuracy and best model parameters are replaced by current accuracies and model parameters.  
Images ($Img$), features ($F_{Img}$), loss ($Loss$), predictions from head ($P_H$), head layer weights ($NN_H$), etc are different for different phases. We wrote them with the same notation for simplicity.

\begin{algorithm}
 \caption{Multimodal Training and Validation}
 \begin{algorithmic}[1]\label{ALG_Train}
 \renewcommand{\algorithmicrequire}{\textbf{Input:}}
 \renewcommand{\algorithmicensure}{\textbf{Output:}}
 \REQUIRE Dataset, Pre-trained Model
 \ENSURE  Trained Model Head ($NN_H$)
 \\ \textit{Initialization} : 
  \\ $NN \leftarrow$ Pre-trained Model
  \\ $NN_H \leftarrow$ New Multimodal Head of Model
  \\ $E_{N} \leftarrow$ Number of Epoch 
  \\ $B_N \leftarrow$ Number of Batch
  \\ $Img[j] \leftarrow$ Images of $j^{th}$ Batch
  \\ $F_{Img}[j] \leftarrow$ Features from $Img[j]$
  \\ $F_{Comb}[j] \leftarrow$ Combined Features
  \\ $Info[j] \leftarrow$ Other Input Information for $j^{th}$ Batch
  \\ $TexLabel[j] \leftarrow$ Text Containing Labels for $j^{th}$ Batch
  \\ $Label[j,D] \leftarrow$ Labels for $j^{th}$ Batch based on Detection
  \\ $P_H[i,j] \leftarrow$ Prediction on $j^{th}$ Batch and $i^{th}$ Epoch 
  \\ $Loss[i,j] \leftarrow$ Loss on $j^{th}$ Batch and $i^{th}$ Epoch 
  \\ $Loss[i] \leftarrow$ Loss on $i^{th}$ Epoch 
  \\ $Acc[i,j] \leftarrow$ Accuracy on $j^{th}$ Batch and $i^{th}$ Epoch 
  \\ $Acc[i] \leftarrow$ Accuracy on $i^{th}$ Epoch
  \\ $Acc\_best \leftarrow$ Best Accuracy
  \STATE Load Pre-trained Model ($NN$)
  \STATE Omit Head Layer of $NN$
  \STATE $Acc\_best$ = 0
  \STATE Save $NN_H$ as the initial best
  \STATE Initialize Optimizer and Scheduler
  \FOR {$i = 1$ to $E_{N}$}
  \FOR {$Phase$ in [`Training', `Validation']}
  \FOR {$j = 1$ to $B_{N}$}
  \STATE Load $Img[j]$, $TexLabel[j]$.
  \STATE Obtain $Info[j]$ from texts.
  \STATE Extract $Label[j,D]$ from $TexLabel[j]$
  \STATE Augment $Img[j]$
  \STATE Extract $F_{Img}[j]$ by applying $NN$ to $Img[j]$
  \STATE Pre-process and normalize $Info[j]$
  \STATE Get $F_{Comb}[j]$ by combining $F_{Img}[j]$ and $Info[j]$
  \STATE Get $P_H[i,j]$ by applying $NN_H$ to $F_{Comb}[j]$
  \STATE Get $Loss[i,j]$ from $P_H[i,j]$ and $Label[j,D]$
  \STATE Get $Acc[i,j]$ from $P_H[i,j]$ and $Label[j,D]$
  \IF{$Phase$ = `Training'}
  \STATE Optimization step based on $Loss[i,j]$
  \ELSE
  \STATE Save $Acc[i,j]$
  \ENDIF
  \ENDFOR
  \IF{$Phase$ = `Validation'}
  \STATE Get $Acc[i]$ from $Acc[i,j]$ values
  \IF{$Acc\_best < Acc[i]$}
    \STATE $Acc\_best = Acc[i]$
    \STATE Save $NN_H$
  \ELSE
    \STATE Load Previous $NN_H$
  \ENDIF
  \ENDIF
  \ENDFOR
  \ENDFOR
 \RETURN $NN_H$ 
 \end{algorithmic} 
 \end{algorithm}

\section{Results}
To validate our proposal, we apply \emph{ResNet-50} and \emph{ViT-L/16} models on Food-101 Multimodal and the FPU23 dataset. Details of the Dataloader and training process are available in Section \ref{proposed}. We apply the same learning rate, momentum, step size, and gamma values for unimodal image and multimodal training combinations. We set the learning rate to $5 \times 10^{-4}$. The value of momentum becomes 0.9, the value of step size becomes 7, and the value of gamma becomes 0.1. The text classification model training on UPMC Food-101 data has different parameter combinations due to the nature of the data and models.

\label{results}

\subsection{UPMC Food-101 Multimodal Classification}
Each sample of the UPMC Food-101 dataset contains an image, a title, and a label. As labels are words, we assign class numbers to labels. As titles are sentences, we checked titles for coherency. All titles contained strings of non-zero length. We observe only one training sample with a single-character string title. We discard that training sample for the text-only and multimodal training. We apply a built-in tokenizer from the \emph{torchtext} library. We also develop a vocabulary of tokens. We encode titles based on the vocabulary and tokenizer. We train a shallow NN of two hidden layers of two hundred neurons each with encoded titles. We keep batch size 128, learning rate 0.001, and epoch number 10. We apply the Adam optimizer with the cross-entropy loss.

Table \ref{TAB_food} presents test accuracies of trained models on the UPMC Food-101 dataset. The unimodal text model provides 83.43\% accuracy on average. We train the image unimodal model with the common training procedure stated at the beginning of this section. \emph{ResNet-50} receives about 59\% accuracy where the \emph{ViT} model receives about 76\% accuracy on average. GPT-2 outperforms the Shallow NN in terms of accuracy. The average accuracy from the GPT-2 model is roughly 1.5\% higher.

Multimodal model provides better accuracy compared to their unimodal parts. The proposed multimodal model training with batch augmentation and unimodal fine-tuning of initial layers brings superior performance. Also, the \emph{ViT-L/16} model performs better than the \emph{ResNet-50} model. According to our literature search, we have received near state-of-the-art (SOTA) performance on the UPMC Food-101 multimodal dataset. We receive 92.63\% accuracy on average with the vision transformer and a shallow NN pair. The SOTA result is 93.1\% \cite{suresh2024stacking}. However, they achieve that result by assembling several models. We achieve the new SOTA performance with the vision transformer and GPT-2 pair. We receive 93.29\% accuracy from that pair on average.

\begin{table*} 
		\caption{Test accuracy through different models and methods on UPMC Food-101 data. }
        \setlength{\tabcolsep}{5pt}
		\label{TAB_food}
		\centering
			\begin{tabular}{|c|c|c|c|c|c|}	\hline
Vision   &Text     & \multicolumn{4}{c|}{Accuracy (\%)}\\ \cline{3-6}
Model     &Model     & Image-only & Text-only & Multimodal& Proposed \\ \hline
\emph{ResNet-50} &Shallow NN     & 59.03$\pm$1.23 & 83.43$\pm$0.33 & 85.60$\pm$0.97 & 89.72$\pm$0.27\\   \hline
\emph{ViT-L/16}  &Shallow NN     & 76.05$\pm$1.47 & 83.43$\pm$0.33 & 91.27$\pm$0.35 & \textbf{92.63$\pm$0.24}\\   \hline

\emph{ResNet-50} &GPT-2 & 59.03$\pm$1.23 & 84.88$\pm$0.38 & 85.71$\pm$0.95 & 91.01$\pm$0.30\\   \hline
\emph{ViT-L/16}  &GPT-2 & 76.05$\pm$1.47 & 84.88$\pm$0.38 & 92.05$\pm$0.41 & 93.29$\pm$0.33\\   \hline

	    \end{tabular} 
\end{table*}	

\subsection{FPU23: Head Detection}
 Table \ref{TAB_ACC} presents the test accuracies for different detection problems, training, and model combinations for the FPU23 dataset. We write a common Dataloader script for all detection. To prepare labels for head detection, we search for the word `Head' on the label containing text. Samples, where the `Head' word is found are labeled as positive samples. Samples, where the `Head' word is not found are labeled as negative samples. We train \emph{ResNet-50} models over two epochs and we train the \emph{ViT-L/16} model over three epochs. Fig. \ref{c_all}(a) presents the confusion matrix on the test subset for head detection.

The first two rows of Table \ref{TAB_ACC} present the test accuracies for the head detection problem. We investigate both models with image-only, multimodal, and proposed multimodal training combinations. According to values of accuracies, proposed training with the \emph{ViT-L/16} model provides the best result. We write the best result among these six combinations as bold text.

\subsection{FPU23: Abdomen Detection}
\begin{figure}
\begin{center}
\includegraphics[clip, trim=2.0cm 2.6cm 8.3cm 1.5cm, width=3.5in,angle=0]{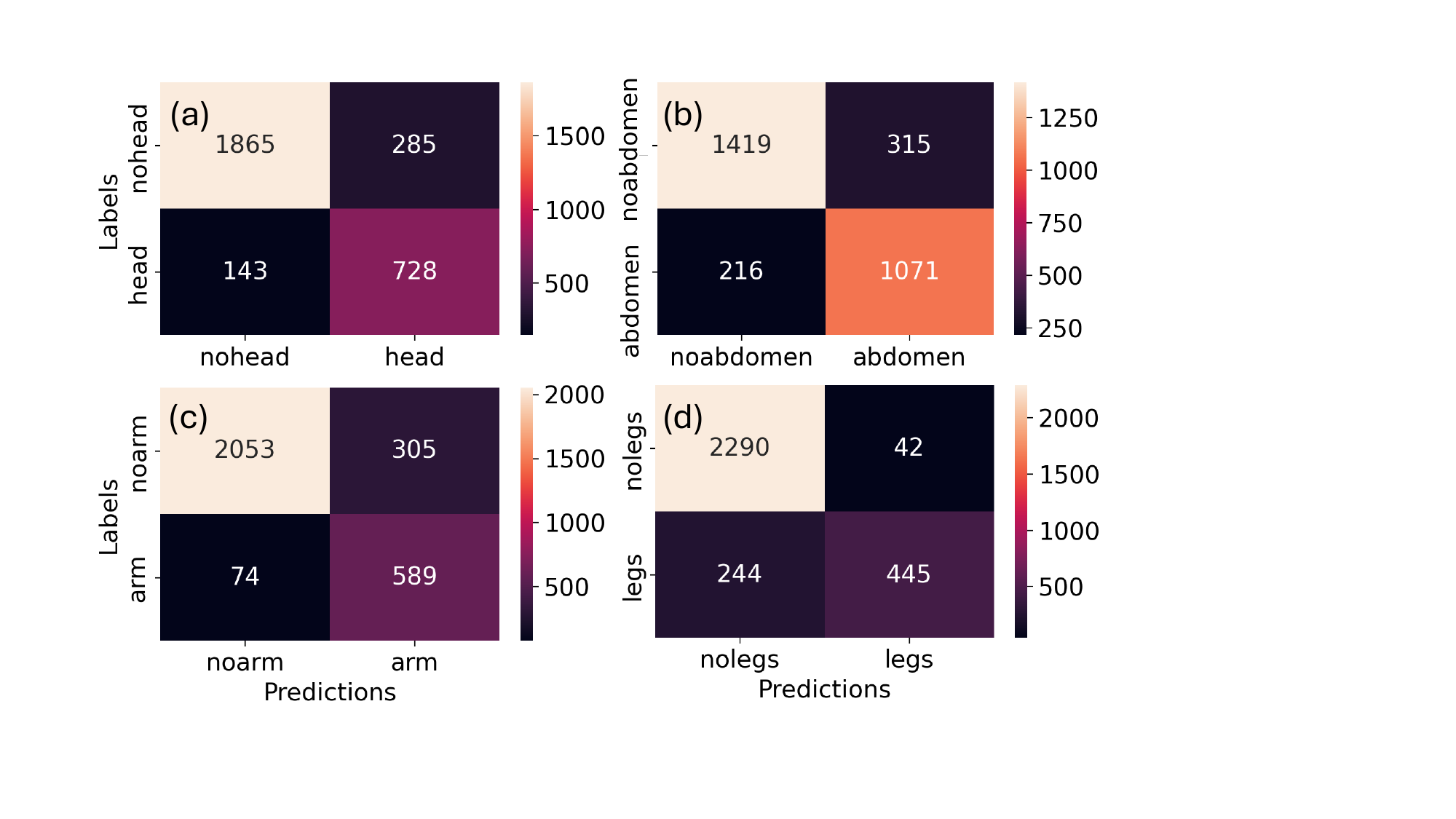}
\caption{Example confusion matrix plots with the multimodal learning on the test set for the (a) `head', (b) `abdomen', (c) `arm', and (d) `leg' detection.}\label{c_all}
\end{center}
\end{figure}

To prepare labels for abdomen detection, we search for the word `Abdomen' on the label containing text. Samples, where the `Abdomen' word is found are labeled as positive samples. Samples, where the `Abdomen' word is not found are labeled as negative samples. Fig. \ref{c_all}(b) presents an example confusion matrix on the test subset for the abdomen detection problem.

We investigate both models with image-only, multimodal, and proposed multimodal training combinations. The third to the fourth rows of Table \ref{TAB_ACC} present the test accuracies for the head detection problem. According to values of accuracies, proposed training with the \emph{ViT-L/16} model provides the best result. We write the best result among these six combinations as bold text.

\subsection{FPU23: Arm Detection}
To prepare labels for arm detection, we search for the word `Arm' on the label containing text. Samples, where the `Arm' word is found are labeled as positive samples. Samples, where the `Arm' word is not found are labeled as negative samples. Fig. \ref{c_all}(c) presents an example confusion matrix on the test subset for the arm detection problem.

The fifth to the sixth rows of Table \ref{TAB_ACC} present the test accuracies for the arm detection problem. According to values of accuracies, image-only training with the \emph{ViT-L/16} model provides the best result. However, the proposed training with \emph{ViT-L/16} provides an accuracy that is very close to the best accuracy. We investigate both models with image-only, multimodal, and proposed multimodal training combinations. We write the best result among these six combinations as bold text. 

\subsection{FPU23: Leg Detection}
To prepare labels for leg detection, we search for the word `legs' on the label containing text. Samples, where the `legs' word is found are labeled as positive samples. Samples, where the `legs' word is not found are labeled as negative samples. Fig. \ref{c_all}(d) presents an example confusion matrix on the test subset for the legs detection problem.

The last two rows of Table \ref{TAB_ACC} present the test accuracies for the leg detection problem. According to values of accuracies, proposed training with the \emph{ViT-L/16} model provides the best result. We write the best result among these six combinations as bold text.

\begin{table} 
		\caption{Test accuracy through different models and methods on Fetal Ultrasound Data. }
		\label{TAB_ACC}
		\centering
			\begin{tabular}{|c|c|c|c|c|}	\hline
    Organ to      &Model  & \multicolumn{3}{c|}{Accuracy (\%)}\\ \cline{3-5}
    Detect   &            & Image-only & Multimodal* & Proposed \\ \hline
    Head     &\emph{ResNet-50} & 71.83$\pm$2.97 & 72.27$\pm$3.53 & 81.41$\pm$1.51\\   \hline
    Head     &\emph{ViT-L/16}  & 83.81$\pm$1.78 & 85.83$\pm$1.69 & \textbf{96.90$\pm$0.45}\\   \hline
    
    Abdomen  &\emph{ResNet-50} & 67.63$\pm$4.03 & 70.57$\pm$3.59 & 80.79$\pm$1.19 \\   \hline
    Abdomen  &\emph{ViT-L/16}  & 80.21$\pm$2.39 & 82.42$\pm$2.26 & \textbf{91.51$\pm$0.79}\\   \hline
    
    Arm      &\emph{ResNet-50} & 75.04$\pm$2.17 & 75.88$\pm$3.23 & 84.16$\pm$1.67\\   \hline
    Arm      &\emph{ViT-L/16}  & 89.08$\pm$2.02 & 88.15$\pm$1.99 & \textbf{93.21$\pm$0.43}\\   \hline
    
    Leg      &\emph{ResNet-50} & 76.69$\pm$3.61 & 77.03$\pm$3.43 & 86.12$\pm$1.76 \\ \hline
    Leg      &\emph{ViT-L/16}  & 91.33$\pm$1.46 & 92.22$\pm$1.53 & \textbf{96.72$\pm$0.34}\\   \hline

	    \end{tabular} 
        \vspace{2pt} \\** The other modality is the viewpoint and fetal orientation information.
\end{table}

\section{Conclusion and Potential Future Work}
\label{conc}
In this paper, we have proposed and demonstrated multimodal learning with batch augmentation, with the fusion of individual unimodal models. We investigate our proposal by organizing labels of ultrasound data to detect images of fetal organs. Also, we investigated the effectiveness of the proposed method on the UPMC Food-101 dataset and achieved state-of-the-art performance. 

Future researchers and developers can potentially apply the proposed method in real-time applications in hospitals and diagnostic centers to serve patients in the future. Researchers can also use the proposed multimodal learning method to predict the age and development of the fetus. Researchers can also use the proposed method for other medical diagnoses and datasets of other modalities. We have extracted certain information from texts during the proposed multimodal method. It is also possible to consider different information based on the available text information. Researchers in the future may also apply our method and our shared scripts to solve similar multimodal problems. Researchers in the future may also develop large datasets of medical images for better initial training on medical data. It is also potentially possible to train an ensemble of models with the proposed method to achieve superior performance.

\bibliographystyle{IEEEtran}
\bibliography{Ref}

\end{document}